\documentclass{article}

    \usepackage[preprint]{neurips_2019}

\usepackage[utf8]{inputenc} 
\usepackage[T1]{fontenc}    
\usepackage{hyperref}       
\usepackage{url}            
\usepackage{amsfonts}       
\usepackage{amsmath}
\usepackage{graphicx}
\usepackage{bbm}
\usepackage{subcaption}
\usepackage{nicefrac}       
\usepackage{microtype}      
\newtheorem{definition}{Definition}

\usepackage{booktabs}       
\usepackage{float}
\usepackage{natbib}

\title{Potential Flow Generator with $L_2$ Optimal Transport Regularity for Generative Models}

\author{%
  Liu Yang and George Em Karniadakis \\
  Division of Applied Mathematics\\
  Brown University\\
  Providence, RI 02906 \\
  \texttt{\{liu\_yang,george\_karniadakis\}@brown.edu} \\
}

\begin{document}

\maketitle

\begin{abstract}
 
We propose a potential flow generator with $L_2$ optimal transport regularity, which can be easily integrated into a wide range of generative models including different versions of GANs and flow-based models. We show the correctness and robustness of the potential flow generator in several 2D problems, and illustrate the concept of ``proximity'' due to the $L_2$ optimal transport regularity. Subsequently, we demonstrate the effectiveness of the potential flow generator in image translation tasks with unpaired training data from the MNIST dataset and the CelebA dataset. 
 
\end{abstract}

\section{Introduction}
Many of the generative models, for example, generative adversarial networks (GANs) \citep{goodfellow2014generative, arjovsky2017wasserstein, salimans2018improving} and flow-based models including normalizing flows \citep{rezende2015variational, kingma2018glow, chen2018neural}, aim to find a generator that could map the input distribution to the target distribution. 

In many cases, especially when the input distribution is purely noises, the specific maps between  input and output are of little importance as long as the generated distributions match the target ones. However, in other cases like image-to-image translations, where both input and target distributions are distributions of images, the generators are required to have additional regularity such that the input individuals are mapped to the ``corresponding'' outputs in some sense. If paired input-output samples are provided, $L_p$ penalty could be hybridized into generators loss functions to encourage the output individuals fit the ground truth \citep{isola2017image}. For the cases without paired data, a popular approach is to introduce another generator and encourage the two generators to be the inverse maps of each other, as in CycleGANs \citep{zhu2017unpaired}, DualGANs \citep{yi2017dualgan} and DiscoGANs \citep{kim2017learning}, etc. However, such a pair of generators is not unique and lacks clear mathematical interpretation about its effectiveness. 

\begin{figure}[ht]
     \centering
         \includegraphics[width=0.5\textwidth]{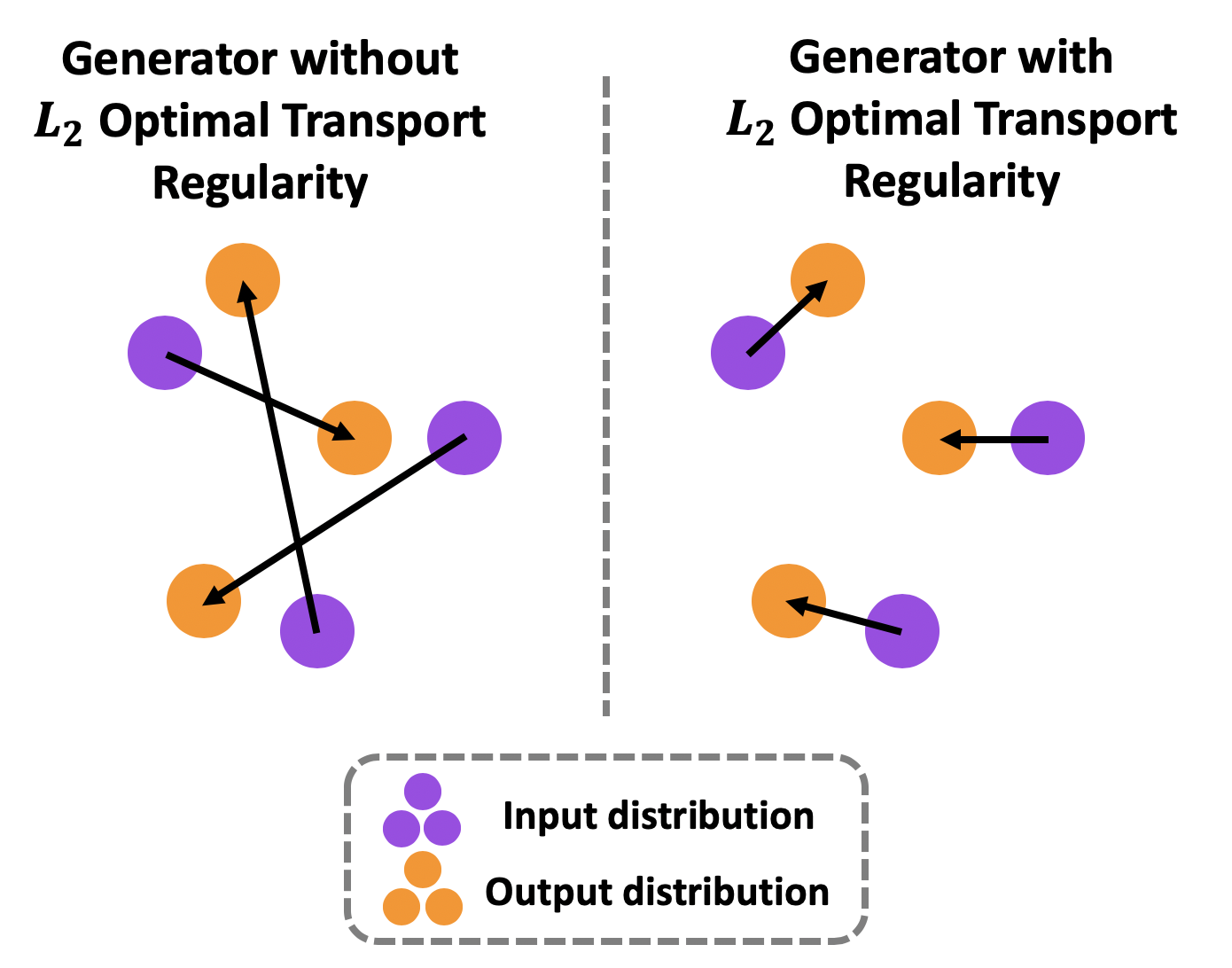}
    \caption{Schematic of generator without and with $L_2$ optimal transport regularity. While both generators provide a scheme to map from the input distribution (purple) to the output distribution (orange), the total squared transport distances in the left generator is much larger than that in the right generator.}
 \label{fig:IntroSchematic}
\end{figure}

In this paper we introduce a special generator, i.e., the potential flow generator, with $L_2$ optimal transport regularity. It is not only a map from the input distribution to the target distribution, but also the {\em optimal transport map} with squared $L_2$ distance as transport cost. In Fig.~\ref{fig:IntroSchematic} we provide a schematic comparison between generators with and without optimal transport regularity. Note that the generator with optimal transport regularity has the characteristic of ``proximity''in that the inputs tend to be mapped to nearby outputs. As we will show later, this ``proximity'' characteristic of $L_2$ optimal transport regularity could be utilized in image translation tasks. Compared with other approaches like CycleGANs, the $L_2$ optimal transport regularity has a much clearer mathematical interpretation.

There have been other approaches to learn the optimal transport map in generative models. 
For example, \cite{seguy2017large} proposed to first learn the regularized optimal transport plan and then the optimal transport map, based on the dual form of regularized optimal transport problem. Also, \cite{yang2018scalable} proposed to learn the unbalanced optimal transport plan in an adversarial way derived from a convex conjugate representation of divergences. In the W2GAN model proposed by \cite{leygonie2019adversarial}, the discriminator’s objective is the 2-Wasserstein metric so that the generator is the $L_2$ optimal transport map. All the above approaches need to introduce, and are limited to, specific loss functions to train the generators. Our proposed potential flow generator takes a different approach in that with up to a slight augmentation of the original generator loss functions, our potential flow generator could be integrated into to a wide range of generative models with various generator loss functions, including different versions of GANs and flow-based models. This simple modification makes our method easy to adopt
on various tasks considering the existing rich literature and the future developments of generative models.

Our main contributions are the following:
\begin{enumerate}
    \item We propose a potential flow generator with $L_2$ optimal transport regularity for a wide range of generative models. Two different versions are proposed: the discrete version and the continuous version.
    \item We compare the two versions and verify the correctness and robustness of the continuous potential flow generator by comparing it with analytical $L_2$ optimal transport maps. We also show the effectiveness of our methods in different generative models for more complicated distributions.
    \item We show the effectiveness of our generator with $L_2$ optimal transport regularity in image translation tasks with unpaired training data.
\end{enumerate}

In Section \ref{sec:GM_and_OTM} we give a formal definition of optimal transport map and the motivation to apply $L_2$ optimal transport regularity to generators. In Section \ref{sec:PFG} we give a detailed formulation of potential flow generator and the augmentation to the original loss functions. Results are  then provided in Section \ref{sec:results}. We include a discussion and conclusion in Section \ref{sec:conclu}.

\section{Generative Models and Optimal Transport Map}\label{sec:GM_and_OTM}

First, we introduce the concept of \textit{push forward}, which will be used extensively in the paper. 
\begin{definition}
Given two Polish space $X$ and $Y$, $B(X)$ and $B(Y)$ the Borel $\sigma$-algebra on $X$ and $Y$, and $\mathcal{P}(X),\mathcal{P}(Y)$ the set of probability measures on $B(X)$ and $B(Y)$. Let $f: X \rightarrow Y$ be a Borel map, and $\mu \in \mathcal{P}(X)$. We define $f_\#\mu \in \mathcal{P}(Y)$, the push forward of $\mu$ through $f$, by
\begin{equation}
    f_\#\mu(E) = \mu(f^{-1}(E)), \forall E \in B(Y).
\end{equation}
\end{definition}

With the concept of push forward, we can formulate the goal of GANs and and flow-based models as to train the generator $G$ such that $G_{\#}\mu$ is equal to or at least close to $\nu$ in some sense, where $\mu$ and $\nu$ are the input and target distribution, respectively. Usually, the loss functions for training the generators are metrics of closeness that vary for different models. For example, in continuous normalizing flows \citep{chen2018neural}, such metric of closeness is $\text{KL}(G_{\#}\mu||\nu)$ or $\text{KL}(\nu||G_{\#}\mu)$. In Wasserstein GANs (WGANs) \citep{arjovsky2017wasserstein}, the metric of closeness is the Wasserstein-1 distance between $G_{\#}\mu$ and $\nu$, which is estimated in a variational form with discriminator neural networks. As a result, the generator and discriminator neural networks are trained in an adversarial way:
\begin{equation}\label{eqn:GAN}
   \min_{G}\max_{D\text{ is 1-Lipschitz}}
                  \mathbb{E}_{x\sim \nu}D(x) - \mathbb{E}_{z\sim \mu}D(G(z)),
\end{equation}
where $D$ is the discriminator neural network and the Lipschitz constraint could be imposed via the gradient penalty \citep{gulrajani2017improved}, spectral normalization \citep{miyato2018spectral}, etc.

Now we can introduce the concept of Monge's optimal transport problem and the optimal transport map as following:

\begin{definition}
Given a cost function $c: X\times Y \rightarrow \mathbb{R}$, and $\mu \in \mathcal{P}(X)$, $\nu \in \mathcal{P}(Y)$, we let $T$ be the set of all transport maps from $\mu$ to $\nu$, i.e. $T := \{f:f_\#\mu = \nu\}$. Monge's optimal transport problem is to minimize the cost functional $C(f)$ among $T$, where 
\begin{equation}
    C(f) = \mathbb{E}_{x\sim \mu} c(x, f(x)) 
\end{equation}
and the minimizer $f^* \in T$ is called the optimal transport map.
\end{definition}

In this paper, we are concerned mostly  about the case where $X = Y =\mathbb{R}^d$ with $L_2$ transport cost, i.e., the transport $c(x, y) = \Vert x - y\Vert_{2}^2$. Also, we assume that $\mu$ and $\nu$ are absolute continuous w.r.t. Lebesgue measure, i.e. they have probability density functions. Such problem is important for the following reasons:

\begin{enumerate}
    \item In general, Monge's problem could be ill-posed in that $T$ could be empty set or there is no minimizer in $T$. Moreover, the optimal transport map could be non-unique, for example, if transport cost $c(x,y) = \Vert x - y\Vert_{1}$. However, for the special case we consider, there exists a unique solution to  Monge's problem. 
    
    \item Informally speaking, with $L_2$ transport cost the optimal transport map has the characteristic of ``proximity'', i.e. the inputs tend to be mapped to nearby outputs. In image translation tasks, such ``proximity'' characteristic would be helpful if we could properly embed the images into Euclidean space such that our preferred input-output pairs are close to each other.
    
    \item Apart from image translations, the optimal transport problem with $L_2$ transport cost is important in many other aspects. For example, it is closely related to gradient flow \citep{ambrosio2008gradient}, Fokker-Planck equations \citep{santambrogio2017euclidean}, porous medium flow \citep{otto1997viscous}, etc. 
\end{enumerate}

\section{Potential Flow Generator}\label{sec:PFG}

\subsection{Potential Flow Formulation of Optimal Transport Map}

One of the most important results for the case we are studying is that the optimal map is of the form $f(x) = \nabla \Phi(x)$, where the so-called Brenier potential $\Phi(x)$ is convex and lower semi-continuous.
We refer the readers to \cite{gangbo1996geometry} and \cite{mccann2011five} for the details. Note that the convexity constraint for $\Phi$ is essential for optimality. Here is an example of non-convex $\Phi$ leading to sub-optimal transport map: let $\mu$ and $\nu$ both be uniform distribution on $[0,1]$, then the identical map which has zero cost is the optimal map, while we can construct a non-convex $\Phi(x) = x-\frac{1}{2}x^2$ which leads to the sub-optimal transport map $f(x) = 1-x$.

In the context of generative models, it looks plausible to represent the Brenier potential $\Phi(x)$ with a neural network and induce the generator from the gradient. However, the constraint of convexity is not trivial for neural networks. Therefore, we turned to another characterization of optimal transport map, i.e., the potential flow formulation which was firstly proposed by \cite{benamou2000computational}: suppose both $\mu$ and $\nu$ admit probability density $\rho_{\mu}$ and $\rho_{\nu}$, and consider all smooth enough density field $\rho(t,x)$ and velocity field $v(t,x)$, where $t \in [0,T]$, subject to the continuity equation as well as initial and final conditions
\begin{equation}\label{eqn:continuity}
\begin{aligned}
    \partial_t \rho + \nabla & \cdot (\rho v) = 0, \\
    \rho(0, \cdot) = \rho_{\mu} &, \quad \rho(T, \cdot) = \rho_{\nu}.
\end{aligned}
\end{equation}
The above equation actually says that such velocity field will induce a transport map: we can construct an ordinary differential equation (ODE)
\begin{equation}\label{eqn:ODE}
    \frac{du}{dt} = v(t, u),
\end{equation}
and the map between initial point to final point gives the transport map from $\mu$ to $\nu$. 

As is proposed by \cite{benamou2000computational}, for the transport cost function $c(x,y) = ||x-y||^2$, the minimal transport cost is equal to the infimum of 
\begin{equation}\label{eqn:W2distance}
    T\int_{\mathbb{R}^d}\int_{0}^T\rho(t,x)|v(t,x)|^2dxdt
\end{equation}
among all $(\rho, v)$ satisfying equation (\ref{eqn:continuity}). The optimality condition is given by
\begin{equation}\label{eqn:potential}
\begin{aligned}
    v(t,x) = \nabla \phi (t,x) \\
    \partial_t \phi + \frac{1}{2} |\nabla \phi |^2 = 0.
\end{aligned}
\end{equation}
In other words, the optimal velocity field is actually induced from a flow with time-dependent potential $\phi(t,x)$.

Note that different from the Brenier potential formulation, there is no explicit requirement for convexity in the potential flow formulation. This helps us to apply this formulation to design our generators with $L_2$ optimal transport regularity.

\subsection{Potential Flow Generator}
The potential $\phi(t,x)$ is the key function to estimate, since the velocity field could be obtained by taking the gradient of the potential and consequently the transport map could be obtained from the ODE Eqn.~\ref{eqn:ODE}. There are two strategies to use neural networks to represent $\phi$. 
\begin{enumerate}
    \item One can take advantage of the fact that the time-dependent potential field $\phi$ is actually uniquely determined by its initial condition from Eqn.~\ref{eqn:potential}, and use a neural network to represent  the initial condition of $\phi$, i.e. $\phi(0,x)$, while approximating $\phi(t,x)$ via time discretization schemes. 
    \item Alternatively, one can use a neural network to represent $\phi(t,x)$ directly and later apply the PDE regularity for $\phi(t,x)$ in Eqn.~\ref{eqn:potential}.
\end{enumerate}

We name the generators defined in the above two approaches as {\em discrete} potential flow generator and {\em continuous} potential flow generator, respectively, and give a detailed formulation as follows.

\subsubsection{Discrete Potential Flow Generator}
In the discrete potential flow generator, we use the neural network $\tilde{\phi}_{0}(x): \mathbb{R}^d \rightarrow \mathbb{R}$ with parameter $\theta$ to represent the initial condition of $\phi(t,x)$, i.e. $\phi(0,x)$. The potential field $\phi(t, x)$ as well as the velocity field $v(t,x)$ could then be approximated by different time discretization schemes, e.g. Eular schemes, Runge-Kutta schemes, etc. As an example, here we use the first-order forward Eular scheme for the simplicity of implementation. To be specific, suppose the time discretization step is $\Delta t$ and the number of total steps is $n$ with $n\Delta t= T$, then for $i = 0,1...n$, $\phi(i\Delta t,x)$ could be represented by $\tilde{\phi}_i(x)$, where
\begin{equation} \label{eqn:phi}
    \begin{aligned}
        \tilde{\phi}_{i+1}(x) = \tilde{\phi}_i(x) -& \frac{\Delta t}{2}|\nabla \tilde{\phi}_i(x) |^2 , \quad \text{for $i = 0,1,2...,n-1$}.
    \end{aligned}
\end{equation}
Consequently, the velocity field $v(i\Delta t,x)$ could be represented by $\tilde{v}_i(x)$, where
\begin{equation} \label{eqn:v}
    \tilde{v}_i(x) = \nabla \tilde{\phi}_i(x), \quad \text{for $i = 0,1...n$}
\end{equation}
Finally, we can build the transport map from Eqs. \ref{eqn:ODE}:
\begin{equation} \label{eqn:f}
\begin{aligned}
    &\tilde{f}_0(x) = x\\
    \tilde{f}_{i+1}(x) = \tilde{f}_i(x) + &\Delta t \tilde{v}_i(\tilde{f}_i(x)), \quad \text{for $i = 0,1,2...n-1$}
\end{aligned}
\end{equation}
with $G(\cdot) = \tilde{f}_n(\cdot)$ be our transport map. The gradient w.r.t. $x$ in Eqns. \ref{eqn:phi}, \ref{eqn:v}, \ref{eqn:f} are realized by automatic differentiation.

The discrete potential flow generator has built-in optimal transport regularity since the optimal condition (Eqn.~\ref{eqn:potential}) is encoded in the time discretization (Eqn.~\ref{eqn:phi}). However, such discretization also introduces nested gradients,  which dramatically increases computational cost when the number of total steps $n$ is increased. In our examples, we set $T = 1$ and $n = 4$.

\subsubsection{Continuous Potential Flow Generator}
In the continuous potential flow generator, we use the neural network $\tilde{\phi}(t, x): \mathbb{R}^{d+1} \rightarrow \mathbb{R}$ with parameters $\theta$ to represent $\phi(t,x)$. Consequently, we can represent the velocity field $v(t,x)$ by $\tilde{v}(t, x)$, where
\begin{equation} \label{eqn:cv}
    \tilde{v}(t,x) = \nabla \tilde{\phi}(t, x).
\end{equation}
With the velocity field we could estimate the transport map by solving the ODE (Eqn.~\ref{eqn:ODE}) using any numerical ODE solver. As an example, we can use the first-order  forward Eular scheme, i.e.
\begin{equation} \label{eqn:cf}
\begin{aligned}
    &\tilde{f}(0,x) = x\\
    \tilde{f}((i+1)\Delta t,x) = \tilde{f}(i\Delta t, x) + &\Delta t \tilde{v}(i \Delta t,\tilde{f}(i\Delta t, x)), \text{for $i = 0,1,2...n-1$}
\end{aligned}
\end{equation}
with $G(\cdot) = \tilde{f}(T ,\cdot)$ be our transport map, where $\Delta t$ is the time discretization step and $n$ is the number of total steps with $n\Delta t= T$. 

In the continuous potential flow generator, increasing the number of total steps would not introduce high order differentiations, therefore we could have very small time steps for a better precision of ODE solver. In practice, we set $T =1 $, $n =10$ in image translation tasks and $n=100$ elsewhere for continuous potential flow generators. Different from the discrete potential flow generator, the optimal condition (Eqn.~\ref{eqn:potential}) is not encoded in the continuous potential flow generator, therefore we need to penalize  Eqn.~\ref{eqn:potential} in the loss function, as we will discuss in the next subsection. 

One may come up with another strategy of imposing the $L_2$ optimal transport regularity: to use a vanilla generator, which is a neural network directly mapping from inputs to outputs, and penalize  the $L_2$ transport cost, i.e., the loss function is 
\begin{equation}\label{eqn:Regloss} 
L_{vanilla} = L_{original} + \alpha \mathbb{E}_{x\sim \mu}\Vert G(x)-x \Vert^2,
\end{equation}
where $L_{original}$ is the original loss function for the generator, and $\alpha$ is the weight for the transport penalty. We emphasize that such strategy is much inferior to penalizing the PDE (Eqn.~\ref{eqn:potential}) in the continuous potential flow generator. When training the vanilla generator with $L_2$ transport penalty, no matter how we weight the $L_2$ transport cost penalty, in principle we always have to make a trade off between ``matching the generated distribution with the target one'' and ``reducing the transport cost'' since there is always a conflict between them, and consequently $G_{\#}\mu$ will be biased towards $\mu$. On the other hand, there is no conflict between matching the distributions and penalizing  Eqn.~\ref{eqn:potential} in the continuous potential flow generator. As a consequence, the continuous potential flow generator is robust with respect to different weights for the PDE penalty. We will show this in Section \ref{sec:results}.

\subsection{Training the Potential Flow Generator}

While the optimal condition (Eqn.~\ref{eqn:potential}) has been considered in the above two generators, the constraints of initial and final conditions are so far neglected. Actually, the constraint of initial and final conditions gives the principle to train the neural network: we need to tune the parameter $\theta$ in the neural network so that $G _\# \mu$ matches $\nu$. This could be done by GANs or flow-based models.

\subsubsection{Loss in GAN models}
In the cases where the available data are samples from $\mu$ and $\nu$, we could apply GAN models to train the potential flow generator. In particular, we replace the generator in GANs with the potential flow generator $G$, feed samples from $\mu$ into $G$, and view the outputs as ``fake data'' of $\nu$. Both the discrete and continuous potential flow generator could be applied to such cases. 

For the discrete potential flow generator, since the optimal transport regularity is already built in, the loss for training $G$ is simply the GAN loss for the  generator, i.e.
\begin{equation} \label{eqn:dloss}
\begin{aligned}
    L_{dPFG} = L_{GAN},
\end{aligned}
\end{equation}
where $L_{GAN}$ actually depends on the specific version of GANs. For example, if we use WGANs with gradient pernalty, then
\begin{equation} \label{eqn:WGANloss}
\begin{aligned}
    L_{GAN} = -\mathbb{E}_{z\sim \mu}D(G(z),
\end{aligned}
\end{equation}
where $D$ is the discriminator neural network.

For the continuous potential flow generator, as mentioned above, we also need to penalize the PDE (Eqn.~\ref{eqn:potential}) for the optimal transport regularity. Inspired by the applications of neural networks in solving forward and backward problems of PDEs, as in PINNs \citep{raissi2017physics1, raissi2017physics2}, we penalize  the squared residual of the PDE on the so-called ``residual" points. In particular, the loss for continuous potential flow generator would be
\begin{equation} \label{eqn:closs}
\begin{aligned}
    L_{cPFG} = L_{GAN} + \lambda \frac{1}{N}\sum_{i=1}^N[\partial_t \tilde{\phi}(t_i,x_i) + \frac{1}{2}|\nabla \tilde{\phi}(t_i, x_i)|^2] ^2,
\end{aligned}
\end{equation}
where $\{(t_i, x_i)\}_{i=1}^N$ are the residual points for the estimating the residual of PDE (Eqn.~\ref{eqn:potential}), and $\lambda$ is the weight for the PDE penalty. While there could be other strategies to select the residual points, here we set them as the points on ``trajectories'' of input samples, i.e.

\begin{equation} \label{eqn:coll}
\begin{aligned}
    \{(t_i, x_i)\}_{i=1}^N = \bigcup_{i=0}^n \bigcup_{x_j \in B} \{( i\Delta t,\tilde{f}(i\Delta t,x_j))\},
\end{aligned}
\end{equation}
where $B$ is the set of batch samples from $\mu$. Note that the coordinates of the residual points involves $\tilde{f}$, but this should not be taken into consideration when calculating the gradient of loss function w.r.t. the generator parameters. 

\subsubsection{Loss in flow-based models}
In the cases where we have density of distributions available, we could apply the continuous potential flow generator in flow-based models. Actually, our continuous potential flow generator perfectly matches the continuous normalizing flow model proposed by by \cite{chen2018neural}. Two ways are proposed to train the generator in continuous normalizing flow: density matching and maximum likelihood training. While both ways could be applied to train the continuous potential flow generators, here we take the latter one as an example, where we assume the density of $\mu$ and samples from $\nu$ are available, and we train the generator to maximize $\mathbb{E}_{y\sim \nu}[\log p_{G_{\#}\mu}(y)]$, where $p_{G_{\#}\mu}$ is the density of $G_{\#}\mu$. Note that this is equivalent to minimizing  $\text{KL}(\nu||G_{\#}\mu)$. Consequently, the loss for the  continuous potential flow generator would be:
\begin{equation} \label{eqn:flowloss}
\begin{aligned}
   L_{cPFG} = - \mathbb{E}_{y\sim \nu}[\log p_{G_{\#}\mu}(y)] + \lambda \frac{1}{N}\sum_{i=1}^N[\partial_t \tilde{\phi}(t_i,x_i) + \frac{1}{2}|\nabla \tilde{\phi}(t_i, x_i)|^2] ^2,
\end{aligned}
\end{equation}
where as in the GAN model, $\{(t_i, x_i)\}_{i=1}^N$ are the residual points for estimating the residual of PDE (Eqn.~\ref{eqn:potential}), and $\lambda$ is the weight for the PDE penalty. 

To estimate $\log p_{G_{\#}\mu}(y)$, we have the ODE that connects the probability density at inputs and outputs of the generator:
\begin{equation} \label{eqn:logden}
\begin{aligned}
   \frac{d}{d t}\log(p(\tilde{f}(t,x))) &= - \nabla \cdot \tilde{v}(t, \tilde{f}(t,x))) = -\Delta \tilde{\phi}(t, \tilde{f}(t,x)),\\
\end{aligned}
\end{equation}
for all $x$ in the support of $\mu$, where the initial probability density $p(\tilde{f}(0,x)) = p_{\mu}(x)$ is the density of $\mu$ at input $x$, while the terminal probability density $p(\tilde{f}(T,x)) = p_{G_\#\mu}(G(x))$ is the density of $G_{\#}\mu$ at output $G(x)$.  The expectation in the loss function is taken with $y\sim \nu$; we set $x = G^{-1}(y)$ and estimate $x$ by solving the ODE 
\begin{equation} \label{eqn:inv}
\begin{aligned}
   \frac{dU}{dt} = -\tilde{v}(T-t, U)
\end{aligned}
\end{equation}
with initial condition $U(0)$ as $y = G(x)$ and $U(T)$ as the corresponding $x = G^{-1}(y)$. 
Note that with the maximum likelihood training, the density of $\mu$ could be unnormalized, since multiplications with $p_{\mu}$ would merely lead to a constant difference in the loss function.

In practice, we also need to discretize Eqns.~\ref{eqn:logden} and \ref{eqn:inv} properly when calculating $\log p_{G_{\#}\mu}(y)$. For example, we use the first-order Euler scheme with time step $\Delta t = 0.01$ for total time interval $T = 1.0$. Note that the discrete potential flow cannot be trivially applied in flow-based models since we found that the time step size is too large to calculate the density accurately.

\section{Results}\label{sec:results}
In this section we show the numerical results of potential flow generators in different generative models. In specific, in Section \ref{sec:PFGinGAN} we show the results of potential flow generators applied on 2D problems and image translation tasks with GAN models, while in Section \ref{sec:PFGinFlow} we show the results in a flow-based model. The numerical experiments are based on Python package TensorFlow\citep{tensorflow2015}.

\subsection{Potential Flow Generator in GAN models}\label{sec:PFGinGAN} 
\subsubsection{2D problems}
In this subsection, we apply the potential flow generator to several two dimensional problems where samples from $\mu$ and $\nu$ are available.

We first study the following two problems where we know analytical solutions for the optimal transport maps. 
\begin{enumerate}
    \item In the first problem we assume that both $\mu$ and $\nu$ are Gaussian distributions. 
    Suppose $\mu = \mathcal{N}(m_1, \Sigma_1)$, $\nu = \mathcal{N}(m_2, \Sigma_2)$, then \citep{gelbrich1990formula} the minimum transport cost between $\mu$ and $\nu$ will be 
\begin{equation} \label{eqn:Gaussian}
\Vert m_1-m_2\Vert^2 + \text{Tr}(\Sigma_1 + \Sigma_2 - 2 (\Sigma_1^{1/2} \Sigma_2 \Sigma_1^{1/2})^{1/2}),
\end{equation}
which is also known as the squared Wasserstein-2 distance between two Gaussian distributions. In particular, we consider the case 
\begin{equation} \label{eqn:GaussianDensity}
\begin{aligned}
   \mu = \mathcal{N}(\begin{bmatrix} 0 \\ 0 \end{bmatrix}, \begin{bmatrix} 0.25 & 0 \\ 0 & 1 \end{bmatrix}), \quad \nu &= \mathcal{N}(\begin{bmatrix} 0 \\ 0 \end{bmatrix}, \begin{bmatrix} 1 & 0 \\ 0 & 0.25 \end{bmatrix}). 
\end{aligned}
\end{equation}

In this case we can check that $f((x,y)) = (2x,0.5y) $ is the optimal transport map. 

\item In the second problem we assume that $\mu$ and $\nu$ are concentrated on concentric rings. 
In polar coordinates, suppose that $\mu$ is the distribution with radius $r$ uniformly distributed on $[0.5,1]$, while angular $\theta$ uniformly distributed on $[0,2\pi]$, and suppose $\nu$ is the distribution with radius $r$ uniformly distributed on $[2,2.5]$, while angular $\theta$ uniformly distributed on $[0,2\pi]$. 
In this case we can check that $f((r,\theta)) = (r+1.5, \theta)$ in polar coordinates is the optimal transport map. 
\end{enumerate}

\begin{figure}[ht]
     \centering
    \begin{subfigure}[b]{0.24\textwidth}
         \centering
         \includegraphics[width=\textwidth]{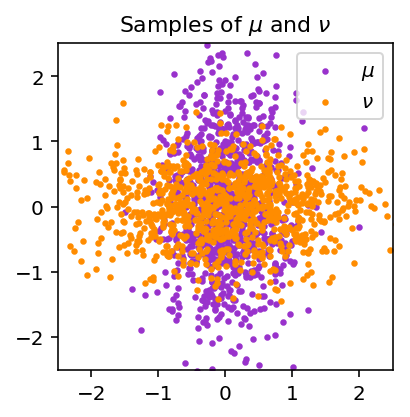}
         \caption{}
         \label{fig:N2Nsamples}
     \end{subfigure}
    \begin{subfigure}[b]{0.24\textwidth}
         \centering
         \includegraphics[width=\textwidth]{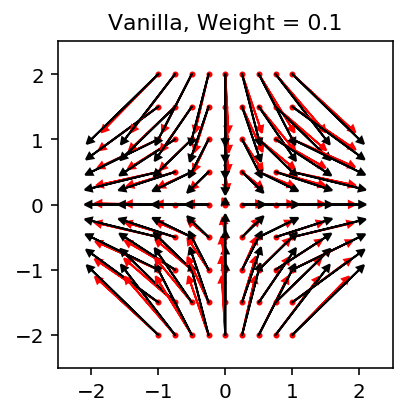}
         \caption{}
         \label{fig:N2NV0.1}
     \end{subfigure}
         \begin{subfigure}[b]{0.24\textwidth}
         \centering
         \includegraphics[width=\textwidth]{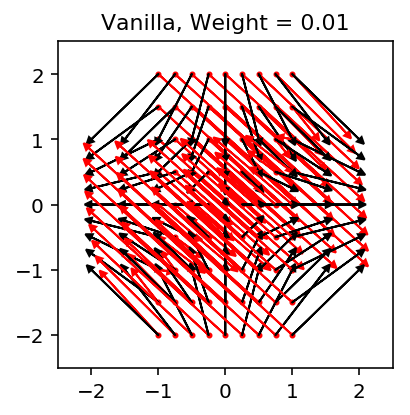}
         \caption{}
         \label{fig:N2NV0.01}
     \end{subfigure}
         \begin{subfigure}[b]{0.24\textwidth}
         \centering
         \includegraphics[width=\textwidth]{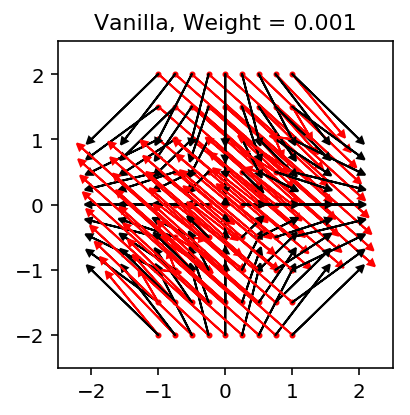}
         \caption{}
         \label{fig:N2NV0.001}
     \end{subfigure}
    \begin{subfigure}[b]{0.24\textwidth}
         \centering
         \includegraphics[width=\textwidth]{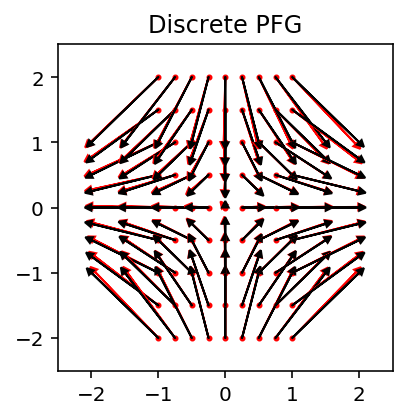}
         \caption{}
         \label{fig:N2ND}
     \end{subfigure}
    \begin{subfigure}[b]{0.24\textwidth}
         \centering
         \includegraphics[width=\textwidth]{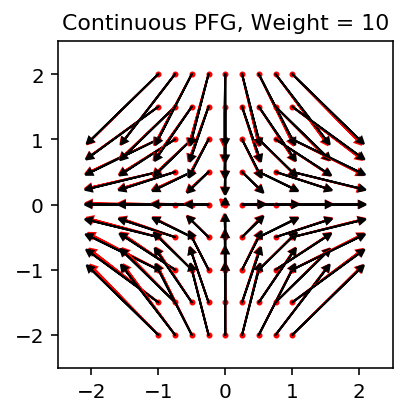}
         \caption{}
         \label{fig:N2NC10}
     \end{subfigure}
         \begin{subfigure}[b]{0.24\textwidth}
         \centering
         \includegraphics[width=\textwidth]{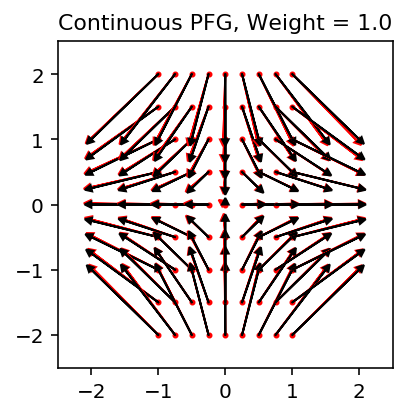}
         \caption{}
         \label{fig:N2NC1}
     \end{subfigure}
         \begin{subfigure}[b]{0.24\textwidth}
         \centering
         \includegraphics[width=\textwidth]{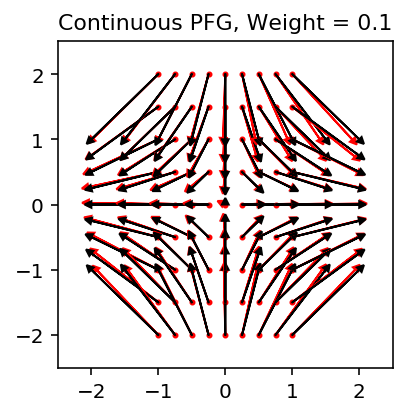}
         \caption{}
         \label{fig:N2NC0.1}
     \end{subfigure}
     
    \caption{Comparison of different methods for problem 1. (a) Samples of $\mu$ (purple) and $\nu$ (orange). (b-d) transport map estimated from the vanilla generators. (e) Discrete potential flow generator (PFG). (f-g) Continuous potential flow generator (PFG). In (b-h), the red arrows represent the estimated transport map while the black arrows represent the reference analytical optimal transport map.}
    \label{fig:N2NSWG}
\end{figure}

\begin{figure}[ht]
     \centering
    \begin{subfigure}[b]{0.24\textwidth}
         \centering
         \includegraphics[width=\textwidth]{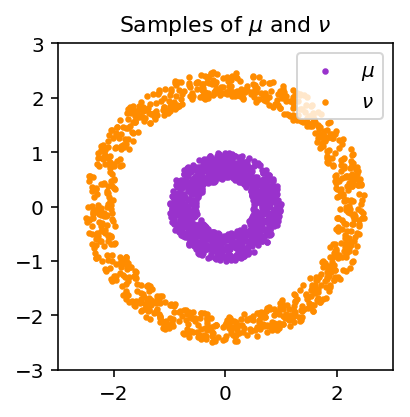}
         \caption{}
         \label{fig:C2Csamples}
     \end{subfigure}
    \begin{subfigure}[b]{0.24\textwidth}
         \centering
         \includegraphics[width=\textwidth]{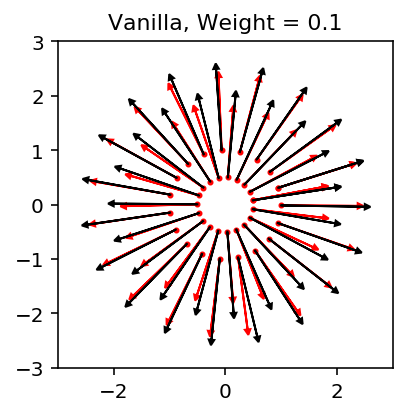}
         \caption{}
         \label{fig:C2CV0.1}
     \end{subfigure}
         \begin{subfigure}[b]{0.24\textwidth}
         \centering
         \includegraphics[width=\textwidth]{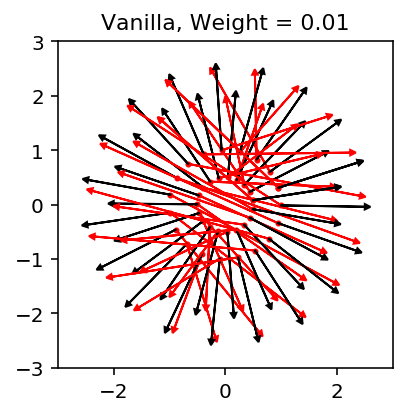}
         \caption{}
         \label{fig:C2CV0.01}
     \end{subfigure}
         \begin{subfigure}[b]{0.24\textwidth}
         \centering
         \includegraphics[width=\textwidth]{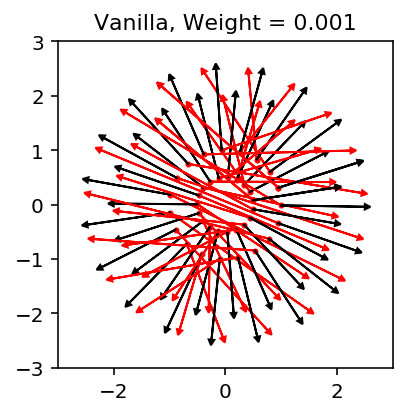}
         \caption{}
         \label{fig:C2CV0.001}
     \end{subfigure}
    \begin{subfigure}[b]{0.24\textwidth}
         \centering
         \includegraphics[width=\textwidth]{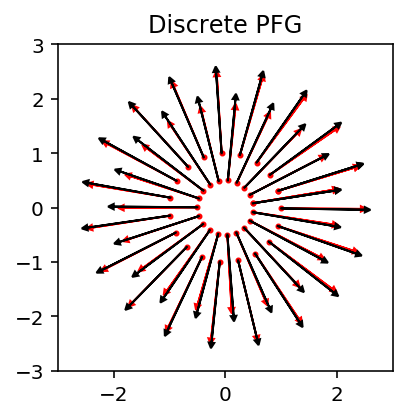}
         \caption{}
         \label{fig:C2CD}
     \end{subfigure}
    \begin{subfigure}[b]{0.24\textwidth}
         \centering
         \includegraphics[width=\textwidth]{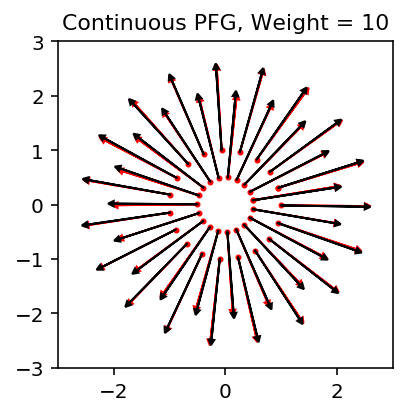}
         \caption{}
         \label{fig:C2CC10}
     \end{subfigure}
         \begin{subfigure}[b]{0.24\textwidth}
         \centering
         \includegraphics[width=\textwidth]{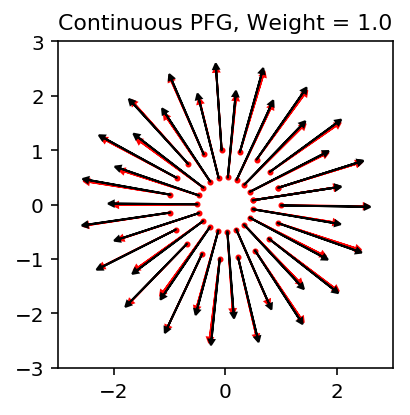}
         \caption{}
         \label{fig:C2CC1}
     \end{subfigure}
         \begin{subfigure}[b]{0.24\textwidth}
         \centering
         \includegraphics[width=\textwidth]{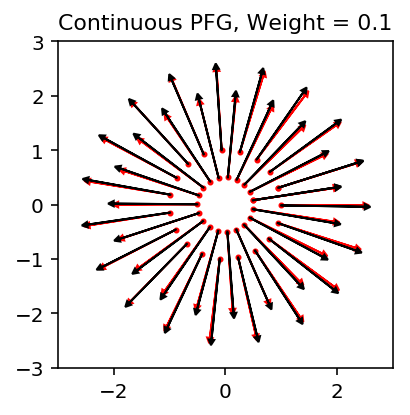}
         \caption{}
         \label{fig:C2CC0.1}
     \end{subfigure}
     
    \caption{Comparison of different methods for problem 2. (a) Samples of $\mu$ (purple) and $\nu$ (orange). (b-d) transport map estimated from the vanilla generators. (e) Discrete potential flow generator. (f-g) Continuous potential flow generator. In (b-h), the red arrows represent the estimated transport map while the black arrows represent the reference analytical optimal transport map.}
    \label{fig:C2CSWG}
\end{figure}

In both cases, we have 40000 samples from each of $\mu$ and $\nu$ as our training data. Samples from $\mu$ and $\nu$ as well as the optimal transport map in both cases are illustrated in Fig.~\ref{fig:N2Nsamples} and Fig.~\ref{fig:C2Csamples}.

For the above two problems we compare the following generators: (a) vanilla generator, i.e., a neural network with a map from $\mathbb{R}^2$ to $\mathbb{R}^2$ that represents the transport map, with $L_2$ transport penalty added to the original loss function, i.e. loss function in Eqn.~\ref{eqn:Regloss} with GAN loss as $L_{original}$, (b) discrete potential flow generator, and (c) continuous potential flow generator with PDE penalty. For the vanilla generator and the continuous potential flow generator, we test different weights for the penalty in order to compare the influences of penalty weights in both methods. For all the generators the neural networks are designed as a feed forward neural network with 5 hidden layers, each of width 128, with a smooth function $\tanh$ as the activation function since the velocity field has to be continuous. We use the Adam optimizer \citep{kingma2014adam} with learning rate $lr = 1e-5$, $\beta_1 = 0.5$, $\beta_2 = 0.9$, and train the neural networks for $100,000$ steps.

As for the GAN loss for generators we use the sliced Wasserstein distance\footnote{Strictly speaking, there is no ``adversarial'' training when we use sliced Wasserstein loss since the distance is estimated explicitly rather than represented by an other neural network. However, the idea of computing the distance between fake data and real data coincide with other GANs, especially WGANs. Therefore, in this paper we view sliced Wasserstein distance as a special version of GAN loss.}, due to its relatively low computational cost, robustness, and clear mathematical interpretation in low dimensional problems \cite{deshpande2018generative}. In particular, we estimate the sliced Wasserstein distances between samples of $G_{\#}\mu$ and $\nu$ using 1000 random projection directions. The batch size is set as 1000. 

For all the cases we run the code three times using different random seed. In Fig.~\ref{fig:N2NSWG} we illustrate the transport maps in one of the three runs for each case. A more systematic and quantitative comparison is provided in Table \ref{tab:N2Ntable} and Table \ref{tab:C2Ctable}, where the best results are marked as bold.

\begin{table}
  \hspace{1em}
  \caption{Problem 1 -- normal to normal distributions.}
  \label{tab:N2Ntable}
  \centering
  \begin{tabular}{llll}
    \toprule
    Generator  & Std in x-axis  & Std in y-axis &  Error of map \\
    \midrule
    Reference &  1.000 & 0.500     \\
    Vanilla ($\alpha = 0.1$) & 0.919$\pm$0.004 & 0.592$\pm$0.003   &  0.108$\pm$0.002\\
    Vanilla ($\alpha = 0.01$)&  0.985$\pm$0.005 & \textbf{0.499$\pm$0.006}   & 0.439$\pm$0.587\\
    Vanilla ($\alpha = 0.001$)&  0.992$\pm$0.009 & 0.493$\pm$0.001   & 0.462$\pm$0.593\\
    Discrete PFG &  \textbf{0.993$\pm$0.001}  & 0.498$\pm$0.002 & \textbf{0.018$\pm$0.006} \\
    Continuous PFG ($\lambda = 10.0$)& 0.991$\pm$0.001 & 0.502$\pm$0.001 & \textbf{0.018$\pm$0.006}  \\
    Continuous PFG ($\lambda = 1.0$)& 0.992$\pm$0.001 & \textbf{0.499$\pm$0.002} & 0.019$\pm$0.007  \\
    Continuous PFG ($\lambda = 0.1$)& 0.990$\pm$0.002 & 0.503$\pm$0.003 & 0.025$\pm$0.008 \\
    \bottomrule
  \end{tabular}
\end{table}

\begin{table}
  \hspace{1em}
  \caption{Problem 2 -- ring to ring distributions.}
  \label{tab:C2Ctable}
  \centering
  \begin{tabular}{lll}
    \toprule
    Generator  & Mean of norm &  Error of map \\
    \midrule
    Reference &  2.250     \\
    Vanilla ($\alpha = 0.1$) & 2.107$\pm$0.002  & 0.146$\pm$0.003    \\
    Vanilla ($\alpha = 0.01$)& 2.227$\pm$0.004 & 0.973$\pm$1.319     \\
    Vanilla ($\alpha = 0.001$)& 2.243$\pm$0.002 & 1.000$\pm$1.311 \\
    Continuous PFG ($\lambda = 10.0$)& 2.243$\pm$0.001 & \textbf{0.024$\pm$0.004} \\
    Continuous PFG ($\lambda = 1.0$)& \textbf{2.245$\pm$0.000} & 0.029$\pm$0.002  \\
    Continuous PFG ($\lambda = 0.1$)& \textbf{2.245$\pm$0.001} & 0.031$\pm$0.004 \\
    \bottomrule
  \end{tabular}
\end{table}

As we already mentioned, the $L_2$ transport penalty would make $G_{\#}\mu$ biased towards $\mu$ to reduce the transport cost from $\mu$ to $G_{\#}\mu$. This is clearly shown in both problems where the penalty weight is $\alpha = 0.1$. Actually, we observed more significant biases with larger penalty weights. For the cases with smaller penalty weights $\alpha = 0.01, 0.001$, in some of the runs, while $G_{\#}\mu$ are close to $\nu$, the estimated transport maps are far from the optimal ones, which shows that the $L_2$ transport penalty cannot provide sufficient regularity if the penalty weight is too small. These numerical results are consistent with our earlier discussion about the intrinsic limitation of the $L_2$ transport penalty.

On the other hand, the potential flow generators give better matching between $G_{\#} \mu$ and $\nu$, as well as smaller errors between the estimated transport maps and the analytical optimal transport maps. Notably, in both problems the continuous potential flow generators give good results with a wide range of PDE penalty weights ranging from $0.1$ to $10$, which shows the superiority of PDE penalty in the continuous potential flow generators compared with the transport penalty in vanilla generators. We also report that while in the first problem the discrete potential flow generator achieves a comparable result with the continuous potential flow generators, in the second problem we encountered ``NAN" problems during training the discrete potential flow generator in some of the runs. This indicates that the discrete potential flow generator is not as robust as the continuous one, which could be attributed to the high order differentiations in the discrete potential flow generators.

\begin{figure}[H]
     \centering
    \begin{subfigure}[b]{0.24\textwidth}
         \centering
         \includegraphics[width=\textwidth]{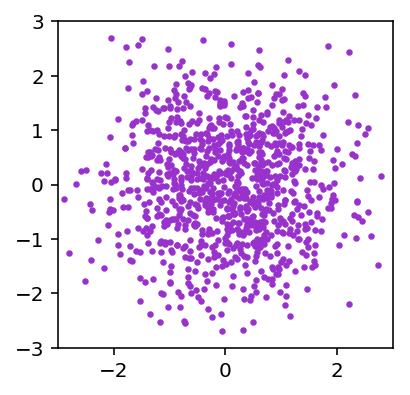}
         \label{fig:128-mu}
     \end{subfigure}
        \begin{subfigure}[b]{0.24\textwidth}
         \centering
         \includegraphics[width=\textwidth]{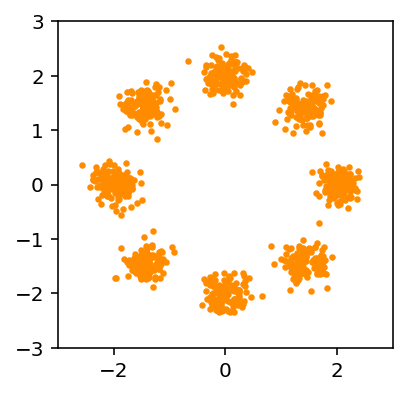}
         \label{fig:128-nu}
     \end{subfigure}
        \begin{subfigure}[b]{0.24\textwidth}
         \centering
         \includegraphics[width=\textwidth]{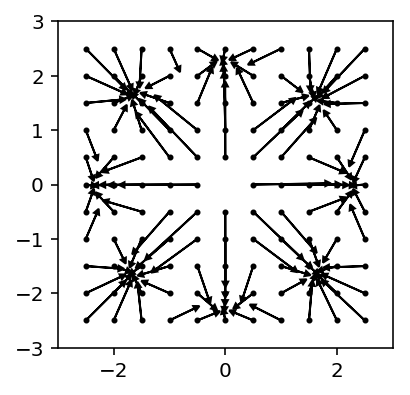}
         \label{fig:128-map}
     \end{subfigure}
         \begin{subfigure}[b]{0.24\textwidth}
         \centering
         \includegraphics[width=\textwidth]{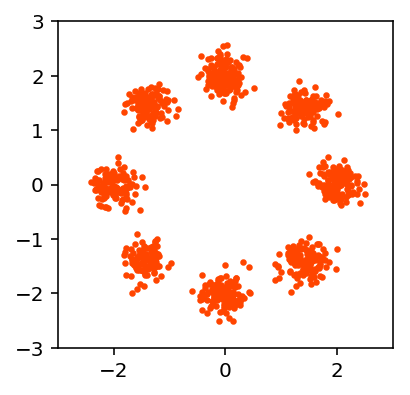}
         \label{fig:128-GP}
     \end{subfigure}
         \begin{subfigure}[b]{0.24\textwidth}
         \centering
         \includegraphics[width=\textwidth]{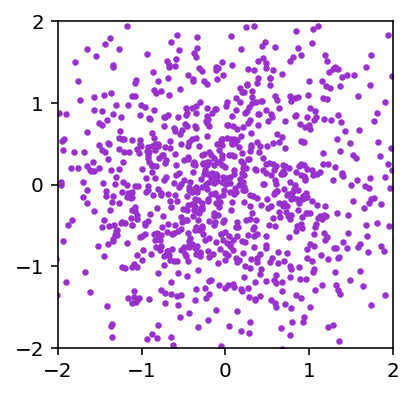}
         \label{fig:N2M-mu}
     \end{subfigure}
        \begin{subfigure}[b]{0.24\textwidth}
         \centering
         \includegraphics[width=\textwidth]{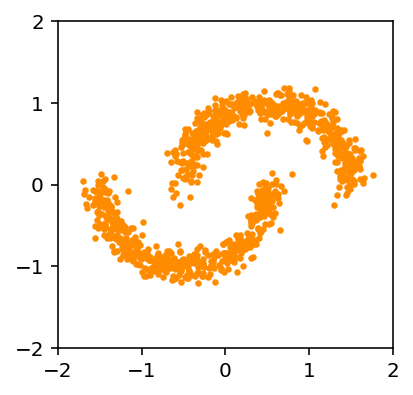}
         \label{fig:N2M-nu}
     \end{subfigure}
        \begin{subfigure}[b]{0.24\textwidth}
         \centering
         \includegraphics[width=\textwidth]{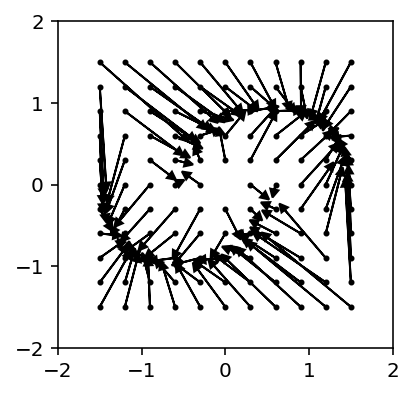}
         \label{fig:N2M-map}
     \end{subfigure}
         \begin{subfigure}[b]{0.24\textwidth}
         \centering
         \includegraphics[width=\textwidth]{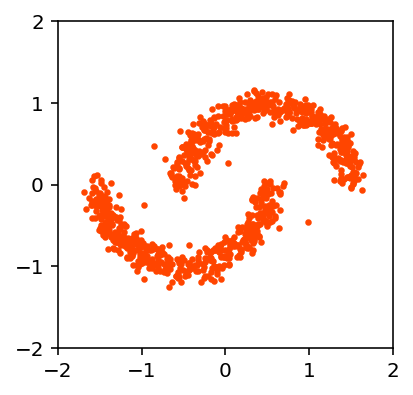}
         \label{fig:N2M-GP}
     \end{subfigure}
         \begin{subfigure}[b]{0.24\textwidth}
         \centering
         \includegraphics[width=\textwidth]{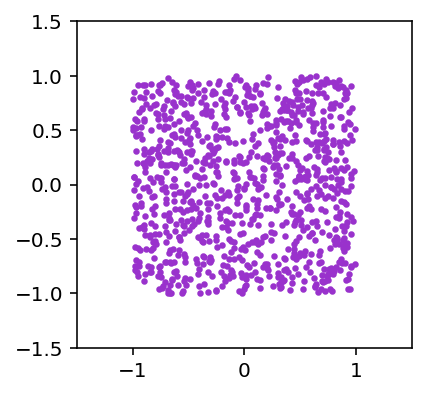}
         \label{fig:S2L-mu}
     \end{subfigure}
        \begin{subfigure}[b]{0.24\textwidth}
         \centering
         \includegraphics[width=\textwidth]{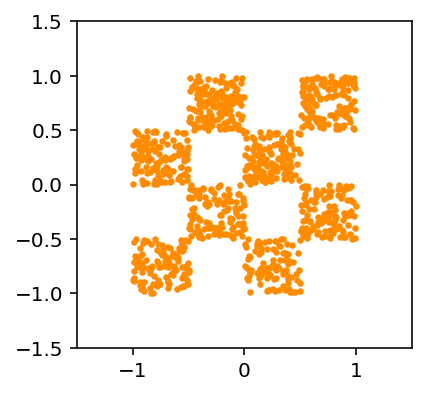}
         \label{fig:S2L-nu}
     \end{subfigure}
        \begin{subfigure}[b]{0.24\textwidth}
         \centering
         \includegraphics[width=\textwidth]{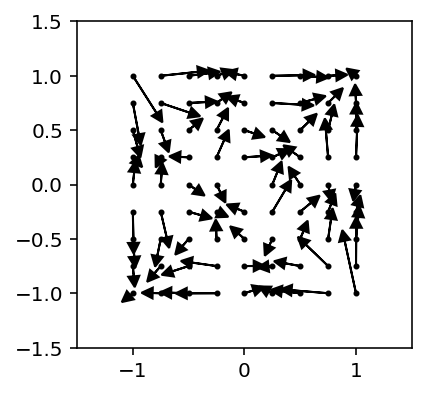}
         \label{fig:S2L-map}
     \end{subfigure}
         \begin{subfigure}[b]{0.24\textwidth}
         \centering
         \includegraphics[width=\textwidth]{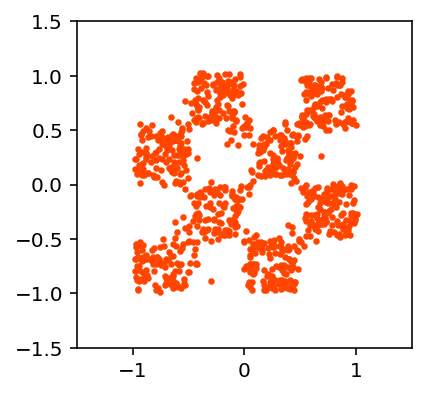}
         \label{fig:S2L-GP}
     \end{subfigure}
    \caption{Continuous potential flow generator in WGAN-GP for three different problems. Each row shows the setup and results of each problem. From left to right: samples of $\mu$, samples of $\nu$, transport map estimated by potential flow generator $G$, samples of $G_{\#}\mu$.}
    \label{fig:2DOther}
\end{figure}

Apart from the previous two problems, we also applied the continuous potential flow generators to another three pairs of $(\mu, \nu)$ distributions, which are illustrated in Fig.~\ref{fig:2DOther}. Here we use WGAN-GP instead of SWG since we found that it provides better GAN loss for complicated distributions. The PDE penalty weight are set as $1.0$.  

From Fig.~\ref{fig:2DOther} we can see the match between $G_{\#}\mu$ and $\nu$ in each of the problems, which shows the effectiveness of the potential flow generator. We could also see that samples of $\mu$ tend to be mapped to nearby positions, which demonstrates the characteristics of ``proximity'' in the potential flow generator maps due to the $L_2$ optimal transport regularity.

\subsubsection{Image Translations: the MNIST and CelebA dataset}

In this section, we aim to show the capability of potential flow generator in dealing with high dimensional problems, and also to show the potential of optimal transport map in tasks of image translations using unpaired training data. In particular, we apply the continuous potential flow generator, due to its robustness compared with the discrete potential flow generator, on the following two tasks: 
\begin{enumerate}
    \item Translation between the MNIST images \citep{mnistdataset}. We divide the MNIST training dataset into two clusters: (a) images of digits 0 to 4, and (b) images of digits 5 to 9. We view the two clusters of images as samples of $\mu$ and $\nu$, respectively, i.e., we want to find the optimal transport map from images of digits 0 to 4 to images of digits 5 to 9.
    \item Translation between the CelebA images \citep{celebAdataset}. We randomly pick 60000 images from the CelebA training dataset and divide them into two clusters: (a) no-smiling face images, i.e., images with attribute ``smiling'' labeled as false, and (b) smiling face images, i.e., images with attribute ``smiling'' labeled as true. We crop the images so that only faces remain on the images, and view the two clusters as samples of $\mu$ and $\nu$, respectively, i.e., we want to find to the optimal transport map from no-smiling face images to smiling face images.
\end{enumerate}
Note that in both tasks there are no paired images in training data, and in the first task we do not distinguish digits among $\mu$ samples and $\nu$ samples.

Before feeding into the generators, we need to embed the images into a Euclidean space, where the $L_2$ distances between embedding vectors should quantify the similarities between images. While there could be other ways for the embedding, like auto-encoders or graph-embedding, in this paper we apply the principal component analysis (PCA)\citep{jolliffe2011principal}, a simple but highly interpretable approach to conduct the image embedding. In both problems, we use WGAN-GP to provide the GAN loss functions. While keeping the time span $T=1.0$, here we set the time step $n=10$ instead of $n = 100$ to reduce the computational cost. The PDE penalty weight $\lambda$is set as 1.0. Both $\tilde{\phi}$ and the discriminator are designed as feed forward neural networks with 5 hidden layers, each of width 256. We emphasize that we did not use any convolutional neural networks here. We use the Adam optimizer with learning rate $lr = 0.0001$, $\beta_1 = 0.5$, $\beta_2 = 0.9$, and train the neural networks for $100,000$ steps.

\begin{figure}[ht]
     \centering
    \begin{subfigure}[b]{1.0\textwidth}
         \centering
         \includegraphics[width=\textwidth]{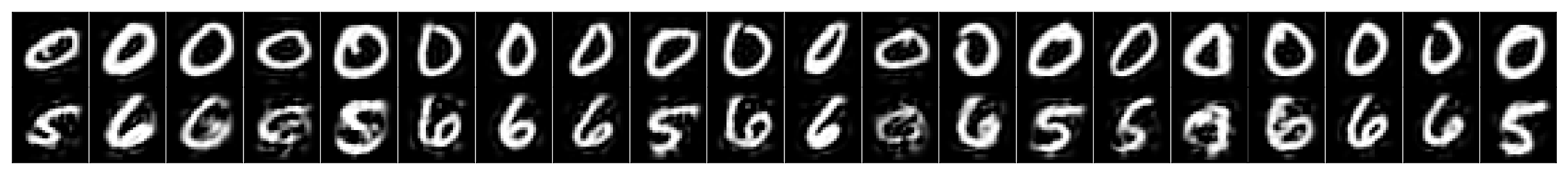}
         \label{fig:MNIST0}
     \end{subfigure}
        \begin{subfigure}[b]{1.0\textwidth}
         \centering
          \vspace{-0.2in}
         \includegraphics[width=\textwidth]{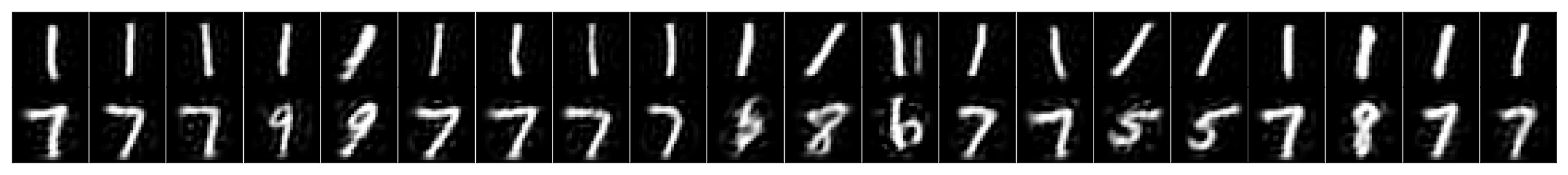}
         \label{fig:MNIST1}
     \end{subfigure}
         \begin{subfigure}[b]{1.0\textwidth}
         \centering
          \vspace{-0.2in}
         \includegraphics[width=\textwidth]{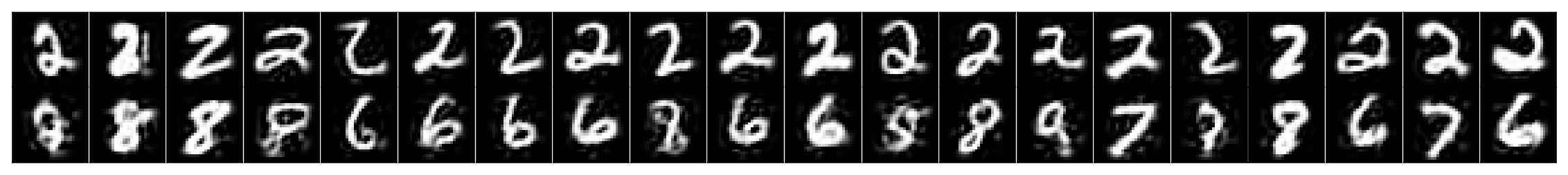}
         \label{fig:MNIST2}
     \end{subfigure}
         \begin{subfigure}[b]{1.0\textwidth}
         \centering
          \vspace{-0.2in}
         \includegraphics[width=\textwidth]{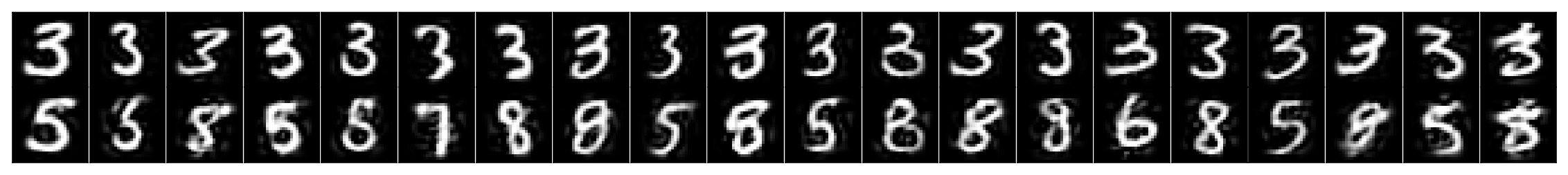}
         \label{fig:MNIST3}
     \end{subfigure}
         \begin{subfigure}[b]{1.0\textwidth}
         \centering
          \vspace{-0.2in}
         \includegraphics[width=\textwidth]{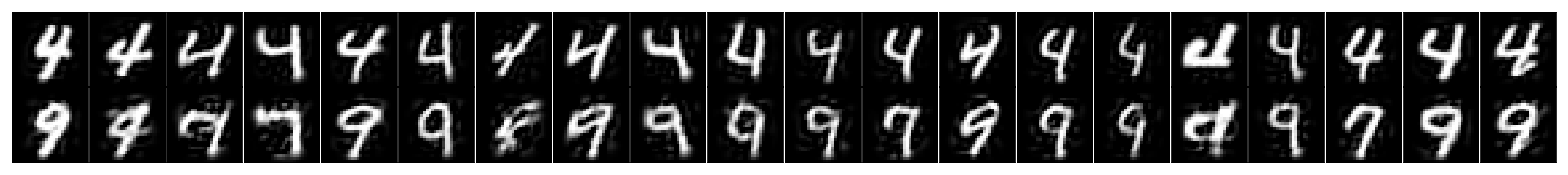}
         \label{fig:MNIST4}
     \end{subfigure}
     
    \caption{Potential flow generator on the MNIST dataset. In each row, the top images are reconstructed from the input vectors, while the bottom images are reconstructed from the corresponding output vectors.}
    \label{fig:MNIST}
\end{figure}

For the first problem on the MNIST dataset, we embed the images into 100-dimensional Euclidean space, i.e. the total components in PCA is 100. We randomly pick images of digits from 0 to 4 from the test set and show the corresponding inputs and outputs in Fig.~\ref{fig:MNIST}. Although the digits are mixed in the training dataset, we can see that after training, the potential flow generator tends to translate images of digit 0 to digit 6, digit 1 to digit 7, digit 3 to digit 5 or 8, and digit 4 to digit 9. This is consistent with our previous discussion about the characteristics of ``proximity'' in that the input digits and output digits ``look similar'', and the corresponding embedding vectors should be close in the $L_2$ distance. 
 
\begin{figure}[H]
     \centering
    \begin{subfigure}[b]{1.0\textwidth}
         \centering
         \includegraphics[width=\textwidth]{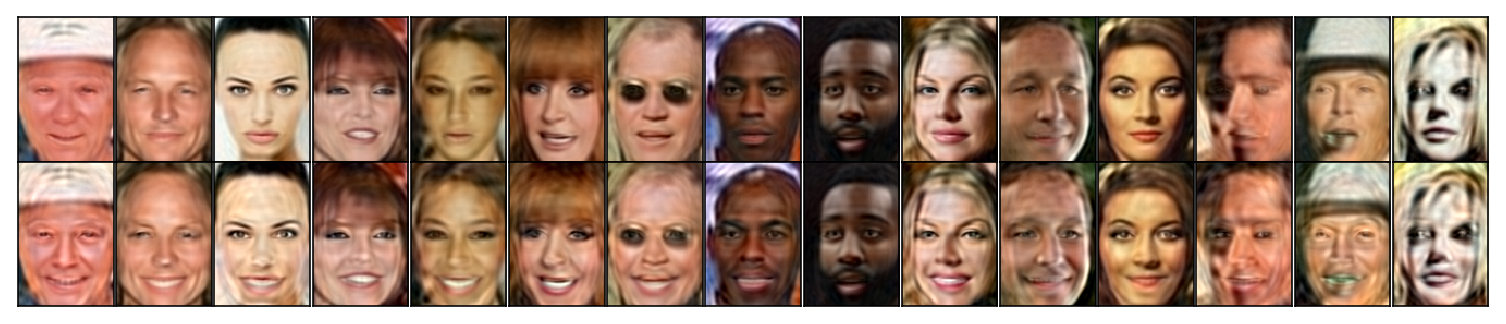}
         \label{fig:CelebA1}
     \end{subfigure}
     \begin{subfigure}[b]{1.0\textwidth}
         \centering
          \vspace{-0.25in}
         \includegraphics[width=\textwidth]{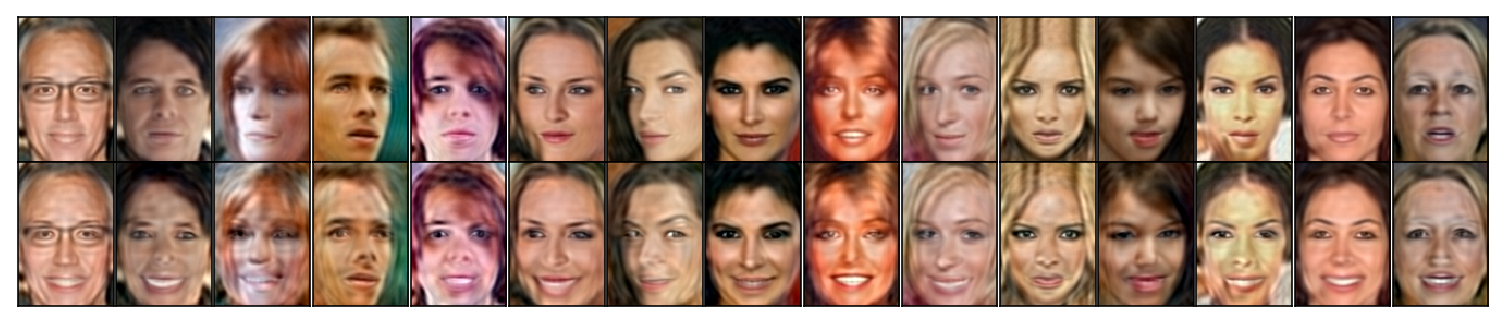}
         \label{fig:CelebA2}
     \end{subfigure}
    \begin{subfigure}[b]{1.0\textwidth}
         \centering
          \vspace{-0.15in}
         \includegraphics[width=\textwidth]{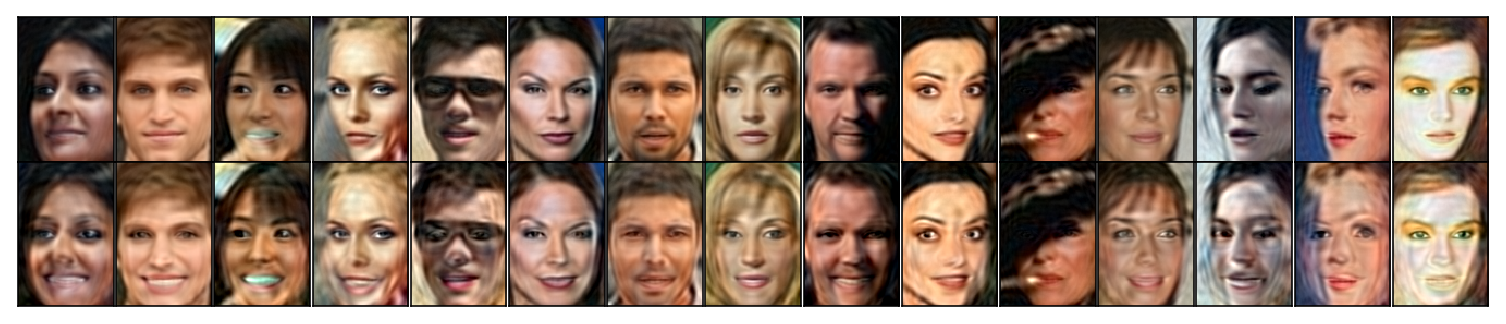}
         \label{fig:CelebA3}
     \end{subfigure}
    \begin{subfigure}[b]{1.0\textwidth}
         \centering
          \vspace{-0.25in}
         \includegraphics[width=\textwidth]{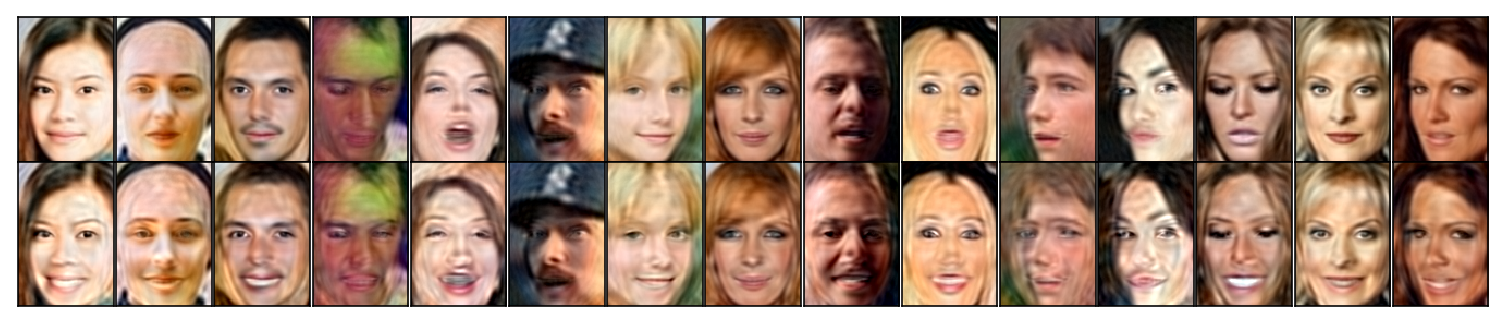}
         \label{fig:CelebA4}
     \end{subfigure}
    \caption{Potential flow generator on the CelebA dataset. In each row, the top images are reconstructed from the input vectors, while the bottom images are reconstructed from the corresponding output vectors. The top two rows are from the training dataset, while the bottom two rows are from the testing dataset. }
    \label{fig:CelebA}
\end{figure}

For the second problem on the CelebA dataset, we embed the images into a 700-dimensional Euclidean space, i.e. the total components in PCA is 700. We randomly pick no-smiling images from the training and testing datasets, and show the corresponding inputs and outputs in Fig.~\ref{fig:CelebA}. We can see that for most of the images, the potential flow generator could translate the no-smiling faces to smiling faces while keeping the other attributes. We can also see the failure for other images, especially for side faces, which could be partially attributed to the fact that these images are outliers in PCA. The reconstructed output images are kind of blurred, since it is difficult to learn the high order modes of PCA.

\subsection{Potential Flow Generator in flow-based models}\label{sec:PFGinFlow}
\begin{figure}[H]
     \centering
    \begin{subfigure}[b]{0.264 \textwidth}
         \centering
         \includegraphics[width=\textwidth]{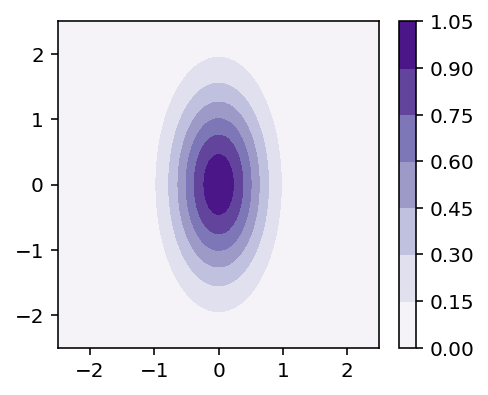}
         \label{fig:Flow-N2N-mu}
     \end{subfigure}
        \begin{subfigure}[b]{0.22\textwidth}
         \centering
         \includegraphics[width=\textwidth]{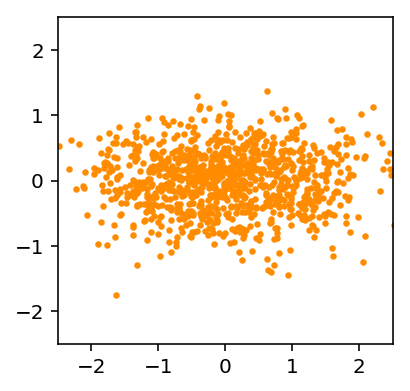}
         \label{fig:Flow-N2N-nu}
     \end{subfigure}
        \begin{subfigure}[b]{0.22\textwidth}
         \centering
         \includegraphics[width=\textwidth]{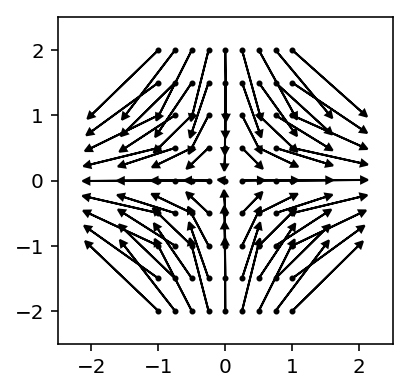}
         \label{fig:Flow-N2N-map}
     \end{subfigure}
         \begin{subfigure}[b]{0.22\textwidth}
         \centering
         \includegraphics[width=\textwidth]{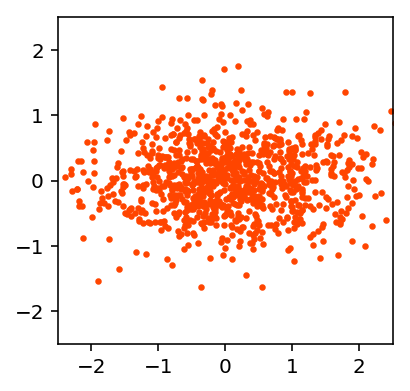}
         \label{fig:Flow-N2N-GP}
     \end{subfigure}
         \begin{subfigure}[b]{0.264 \textwidth}
         \centering
         \includegraphics[width=\textwidth]{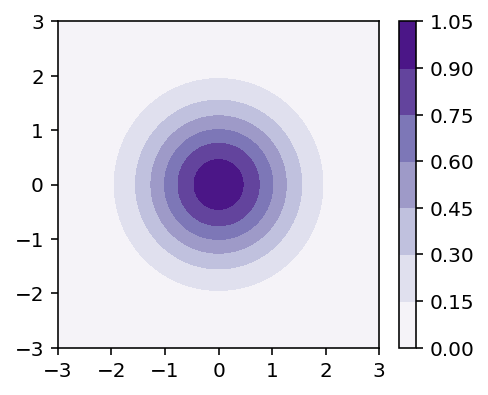}
         \label{fig:Flow-128-mu}
     \end{subfigure}
        \begin{subfigure}[b]{0.22\textwidth}
         \centering
         \includegraphics[width=\textwidth]{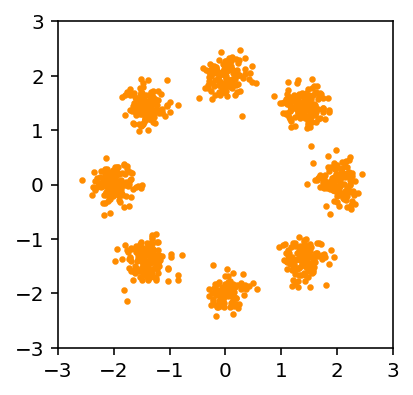}
         \label{fig:Flow-128-nu}
     \end{subfigure}
        \begin{subfigure}[b]{0.22\textwidth}
         \centering
         \includegraphics[width=\textwidth]{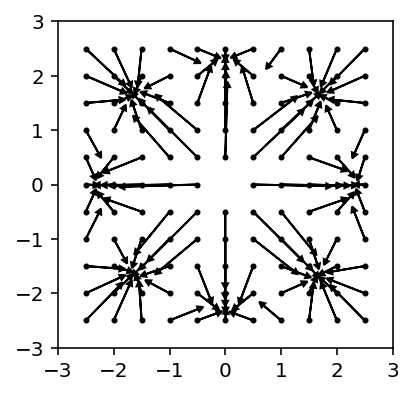}
         \label{fig:Flow-128-map}
     \end{subfigure}
         \begin{subfigure}[b]{0.22\textwidth}
         \centering
         \includegraphics[width=\textwidth]{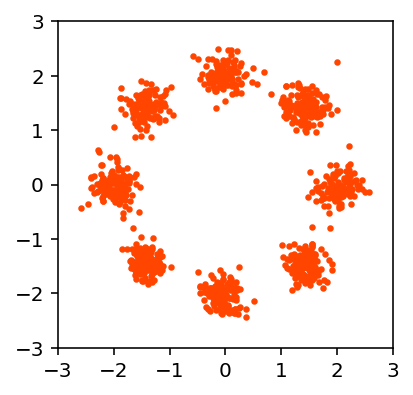}
         \label{fig:Flow-128-GP}
     \end{subfigure}
         \begin{subfigure}[b]{0.264 \textwidth}
         \centering
         \includegraphics[width=\textwidth]{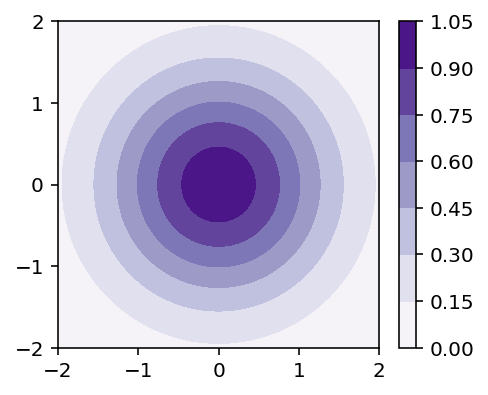}
         \label{fig:Flow-N2M-mu}
     \end{subfigure}
        \begin{subfigure}[b]{0.22\textwidth}
         \centering
         \includegraphics[width=\textwidth]{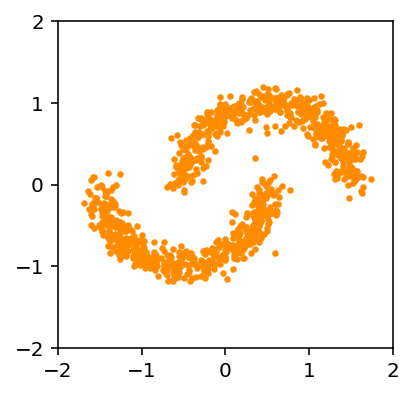}
         \label{fig:Flow-N2M-nu}
     \end{subfigure}
        \begin{subfigure}[b]{0.22\textwidth}
         \centering
         \includegraphics[width=\textwidth]{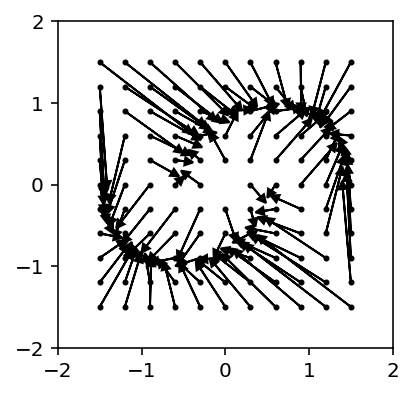}
         \label{fig:Flow-N2M-map}
     \end{subfigure}
         \begin{subfigure}[b]{0.22\textwidth}
         \centering
         \includegraphics[width=\textwidth]{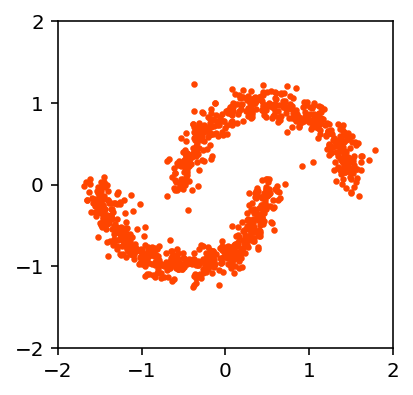}
         \label{fig:Flow-N2M-GP}
     \end{subfigure}
    \caption{Potential flow generator in continuous normalizing flow for three different problems. Each row shows the setup and results of one problem. From left to right: unnormalized density of $\mu$, samples of $\nu$, transport map estimated by potential flow generator $G$, and samples of $G_{\#}\mu$.}
    \label{fig:Flows}
\end{figure}

In this section we apply the continuous potential flow generator in continuous normalizing flows with three pairs of $(\mu, \nu)$ distributions. The distributions of $\mu$ and $\nu$ as well as the results are illustrated in Fig.~\ref{fig:Flows}. We use a feed forward neural network with 5 hidden layers, each of width 128, to represent $\tilde{\phi}$. We use the Adam optimizer with learning rate $lr = 0.0001$, $\beta_1 = 0.9$, $\beta_2 = 0.999$, and train the neural networks for $10,000$ steps. The PDE penalty weights are set as $1.0$. From Fig.~\ref{fig:Flows} we can see the match between $G_{\#}\mu$ and $\nu$ in each of the problems, as well as that that the samples of $\mu$ tend to be mapped to nearby positions, which shows the effectiveness of the continuous potential flow generator in the flow-based models.

\section{Conclusions}\label{sec:conclu}

In this paper we propose potential flow generators with $L_2$ optimal transport regularity. In particular, we propose two versions: the discrete one and the continuous one. Both potential flow generators could be integrated in GAN models with various generator loss functions, while the latter one could also be integrated into flow-based models. For the discrete potential flow generator, the $L_2$ optimal transport is directly encoded in, while for the continuous potential flow generator we only need a slight augmentation to the original generator loss functions to impose the $L_2$ optimal transport regularity.

We first show the correctness of potential flow generators in estimating $L_2$ optimal transport map by comparing with analytical reference solutions. We report that the continuous potential flow generator outperforms the discrete one in robustness. We also illustrated the characteristic of ``proximity'' for potential flow generator due to $L_2$ optimal transport regularity, and consequently the effectiveness of the potential flow generator in image translation tasks using unpaired training data from the MNIST dataset and the CelebA dataset.

Apart from image-to-image translations, it is also possible to apply the potential flow generator to other translation tasks, text-to-text translations for example, if the translation objects could be properly embedded into Euclidean space. It is also interesting to study the application of potential flow generator in other generative models apart from the ones in this paper. Moreover, while in this paper we use PCA for image embedding, a possible improvement is to integrate potential flow generator with other embedding techniques like autoencoders and graph embedding methods. We leave these possible improvements to future work.

\section{Acknowledgement}
This work is supported by the ARO MURI project (W911NF-15-1-0562), the DOE PhILMs project (DE-SC0019453), and the DARPA AIRA project (HR00111990025).

\bibliography{PFG.bib}
\small {\bibliographystyle{plainnat}}

\end{document}